\documentclass[letterpaper, 10 pt, conference]{ieeeconf}

\IEEEoverridecommandlockouts                              

\usepackage[T1]{fontenc}
\usepackage{aecompl}

\usepackage{subcaption}
\usepackage{makecell}
\usepackage{graphicx}
\usepackage{multirow}
\usepackage{amsmath}
\usepackage{amssymb}
\usepackage{booktabs}
\usepackage{algorithm}
\usepackage[noend]{algpseudocode}
\usepackage{cite}
\usepackage[pagebackref,breaklinks,colorlinks]{hyperref}
\usepackage[normalem]{ulem} 

\iffalse 
    \newcommand{\holger}[1]{\noindent}
    \newcommand{\ww}[1]{\noindent}
    \newcommand{\yclin}[1]{\noindent}
\else
    \newcommand{\holger}[1]{[{\bf \color{orange} HC: #1}]}
    \newcommand{\ww}[1]{[{\bf \color{magenta} JW: #1}]}
    \newcommand{\yclin}[1]{[{\bf \color{blue} YL: #1}]}
\fi

\newcommand{\Eq}[1]{Eq.~(\ref{eq:#1})}
\newcommand{\eq}[1]{\Eq{#1}}
\newcommand{\fig}[1]{Fig.~\ref{fig:#1}}
\newcommand{\tab}[1]{Tab.~\ref{tab:#1}}

\newcommand{\sect}[1]{Section~\ref{sec:#1}}

\usepackage{arydshln}

\usepackage{xcolor}
\usepackage{graphicx}
\usepackage{amsmath}
\usepackage{mathtools}  
\usepackage{amsfonts} 
\usepackage{amssymb}
\usepackage{booktabs}
\usepackage{comment}
\usepackage{dblfloatfix} 

\usepackage{xspace}

\title{\LARGE \bf
BaSAL: Size-Balanced Warm Start Active Learning for LiDAR Semantic Segmentation
}

\author{
Jiarong Wei, Yancong Lin$^{\ast}$ and Holger Caesar 
\thanks{$\ast$ Corresponding author. \newline \indent All authors are with the Intelligent Vehicles Group, Delft University of Technology, the Netherlands.}
}%
\date{}
\begin{document}
\maketitle
\thispagestyle{empty}
\pagestyle{empty}

\begin{abstract}
Active learning strives to reduce the need for costly data annotation, by repeatedly querying an annotator to label the most informative samples from a pool of unlabeled data, and then training a model from these samples. 
We identify two problems with existing active learning methods for LiDAR semantic segmentation.
First, they overlook the severe class imbalance inherent in LiDAR semantic segmentation datasets. 
Second, to bootstrap the active learning loop when there is no labeled data available, they train their initial model from randomly selected data samples, leading to low performance. This situation is referred to as the cold start problem.
To address these problems we propose BaSAL, a size-balanced warm start active learning model, based on the observation that each object class has a characteristic size.
By sampling object clusters according to their size, we can thus create a size-balanced dataset that is also more class-balanced.
Furthermore, in contrast to existing information measures like entropy or CoreSet, size-based sampling does not require a pretrained model, thus addressing the cold start problem effectively.
Results show that we are able to improve the performance of the initial model by a large margin. 
Combining warm start and size-balanced sampling with established information measures, our approach achieves comparable performance to training on the entire SemanticKITTI dataset, despite using only 5\% of the annotations, outperforming existing active learning methods. 
We also match the existing state-of-the-art in active learning on nuScenes.
Our code is available at: \url{https://github.com/Tony-WJR/BaSAL}.
\end{abstract}
\section{Introduction}
\PARstart{A}{utonomous} vehicles are often equipped with LiDAR, a time-of-flight sensor that accurately scans the environment and excels at 3D perception tasks.
Semantic segmentation is a popular task for LiDAR-based perception systems whose goal is to assign a predefined class label for each point in the LiDAR scan.
Existing methods on this task are mostly learning-based and evaluated on large-scale benchmark datasets like SemanticKITTI~\cite{SemanticKitti} and nuScenes~\cite{nuScene,fong2022panopticnuscenes}. 
However, training such models requires a large amount of data that is expensive to label. 
Active learning is a machine learning technique that reduces the demand for annotation by interactively selecting informative samples and querying an oracle for annotation~\cite{alliterature}.

Research on active learning for LiDAR (semantic) segmentation~\cite{redal, lidal} focuses on novel information measures that quantify the importance of unlabeled data, such that a limited annotation budget is spent on the most informative samples.
We identify two problems that have not been explicitly addressed in existing active learning methods for LiDAR segmentation: class imbalance and the cold start problem.

The performance on the LiDAR segmentation task is commonly measured using the mean Intersection over Union (mIoU) metric, which assigns the same weight to all classes.
However, since there is a strong class imbalance, models are trained on fewer points of rare classes, thus yielding a lower performance in general.
Consequently, points belonging to rare classes have a relatively larger impact on the mIoU metric. 
In contrast, the active learning budget is measured by the number of points, which treats points of all classes the same.
This reveals a fundamental mismatch between the performance metric and the active learning budget.
By reducing class imbalance in the sampled data, we can reduce this mismatch.

In our initial approach to address class imbalance, we tried to extract more samples of rare classes using the prediction result of pretrained networks. We utilized a model trained on a small subset of labels to classify the unlabeled pool, specifically targeting samples predicted as rare classes. 
Our preliminary studies, however, revealed this strategy to be ineffective due to the model's poor performance and inaccurate predictions on rare classes.
To find a more effective reflection of classes, we observed that objects with identical labels typically share similar sizes. 
Consequently, by creating a size-balanced dataset, we inherently foster a more class-balanced dataset. 

The cold start problem~\cite{nath2022warm} in active learning is about which data to select first for label acquisition and model training when there is no labeled data to start with.
Existing methods initialize the active learning loop by annotating randomly sampled data from a pool of unlabeled data~\cite{lidal, redal} for training. 
Then they select the most informative samples using their proposed information measure based on the outputs of the trained model.
However, the performance of the initial model by random sampling is often lacking, which negatively affects subsequent active learning iterations. 
In contrast, we discovered that sampling objects based on their size does not require a pretrained model. It thus can be easily employed in the very first iteration of active learning, also making the selected data more class-balanced. 
If the initial model trained by this sampling strategy is verified to perform better than the traditional random sampling, it can be a warm start solution to the cold start problem. 

Following the above intuitions, we propose BaSAL, a size-balanced warm start active learning strategy that addresses class imbalance and the cold start problem. 
To address class imbalance, we introduce Average Point Information (API), a metric assessing the informativeness of each object cluster based on its size.
We use the API metric to create partitions of clusters that have approximately the same amount of information and allocate the labeling budget equally to all partitions.
To initiate model training, we randomly draw samples from each partition for label acquisition. 
For subsequent iterations of active learning, we adopt the well-established information measures, entropy~\cite{entropy} and CoreSet~\cite{coreset}, to rank unlabeled data and select top-ranking clusters from each partition for further label acquisition. 
Experiments on SemanticKITTI show that by labeling only $5\%$ points, we are able to achieve a comparable performance of fully supervised learning. 
We also outperform existing active learning methods by a large margin on SemanticKITTI.
On nuScenes, our model performs competitively with the state-of-the-art LiDAL~\cite{lidal} method and reaches $95\%$ performance of fully supervised learning.

In summary, our contribution can be outlined as follows:
\begin{itemize}
    \item Based on the insight that it is non-trivial to sample objects according to their class, we instead propose a method that addresses class imbalance by sampling object clusters according to their size.
    
    \item We initialize the active learning loop by equally distributing the labeling budget to all size-based partitions and then randomly drawing samples from each partition, thus addressing the cold start problem.

    \item Our experiments demonstrate that BaSAL outperforms the state-of-the-art in active learning on SemanticKITTI and matches it on nuScenes.
\end{itemize}
\section{Related Work}
In this section we discuss the literature on active learning, class imbalance and the cold start problem in active learning.
We focus on pool-based active learning that assumes a large pool of unlabeled data available~\cite{alliterature}.

\subsection{Active Learning}
There are numerous strategies to select new samples for label acquisition, including uncertainty-based methods, diversity-based methods or combinations of multiple approaches. 
Uncertainty-based methods select hard examples by measuring model disagreement~\cite{beluch2018power, lakshminarayanan2017simple, tan2024crossmodalityal}, entropy~\cite{entropy, gal2017deep}, predicted loss~\cite{learningloss}, discriminator scores~\cite{variational}, or geometric distances to the decision boundary~\cite{tong2001support}.
However, uncertainty-based methods tend to draw similar samples without taking diversity into account.
Other works~\cite{coreset, preclustering, guo2010active, gudovskiy2020deep} propose a diversity-based strategy that finds a subset of samples that best represents the entire pool. 
Follow-up works~\cite{batchbald, deepbatch, anno3d} combine model uncertainty and diversity. 

Recent research on active learning for LiDAR segmentation takes similar strategies, but exploits task-specific prior knowledge.
LiDAL~\cite{lidal} considers inconsistencies in model predictions across frames as the uncertainty measure for informative sample selection. ReDAL~\cite{redal} divides a point cloud into regions and then selects diverse regions according to multiple cues, including softmax entropy, color discontinuity, and structural complexity.
\cite{liang2021exploring} proposes a diversity-based method for 3D object detection by enforcing both spatial diversity and temporal diversity. \cite{justlabel} uses detection and prediction entropy as the information measures for the prediction and planning tasks. Similarly, we consider both uncertainty and diversity as information measures.
\subsection{Class Imbalance in Active Learning}
Class imbalance is a common problem for datasets collected in the wild.
~\cite{ci_al} summarizes two kinds of techniques to cope with the class imbalance problem: 
Density-sensitive active learning and skew-specialized active learning~\cite{reduce_ci}. The former assigns an informativeness score to each sample by imposing an assumption on the input space. Examples include information density~\cite{seq_label}, pre-clustering~\cite{pre_cluster}, and alternate density-sensitive heuristics~\cite{paired}. 
The latter incorporates a bias towards underrepresented classes, thus resulting in a more balanced sampling~\cite{tomanek2009reducing,ertekin2007learning, bloodgood2014taking }.
\cite{aggarwal2020active} addresses class imbalance in active learning for visual tasks by introducing a sample balancing step that prioritizes minority classes. 
Most methods require a pretrained model. In contrast, we tackle class imbalance in LiDAR segmentation without pretraining by creating size-balanced partitions, as size is a characteristic trait of class.

\subsection{Cold Start in Active Learning} 
Cold start in active learning refers to the problem of which data to label first given an unlabeled pool of data~\cite{nath2022warm}. 
The labeled data is then used to train an initial model which serves as the starting point for active learning.
\cite{coldstartalnlp, a_simple} take advantage of pretrained models and cluster unlabeled samples in the embedding space. 
Those samples closer to the cluster centers are selected for annotation, as they better represent categories than random samples.
\cite{firstchoice} focuses primarily on selecting examples that are hard to learn via self-supervised contrastive learning, based on the assumption that if a model can not separate a sample from others, this sample is expected to exhibit typical characteristics, such as common visual patterns in vision that are shared by others.
\cite{al_on_budget} proposes TypiClust that aims to find out typical examples which better represent the entire dataset. 
The typicality of a sample is measured by the inverse of the average Euclidean distance to its K-nearest neighbors in the feature space.
\cite{covering_lens} proposes ProbCover, a solution for cold start that maximizes the probability of covering the unlabeled set in the embedding space.
\cite{nath2022warm} uses pseudo labels to pretrain models and then rank unlabeled data by uncertainty, after which the top-ranking samples are used for label acquisition. We handle the cold start problem by a simple size-based sampling.
\section{Method}
\begin{figure*}
    \centering
    \includegraphics[width=1.0\textwidth]{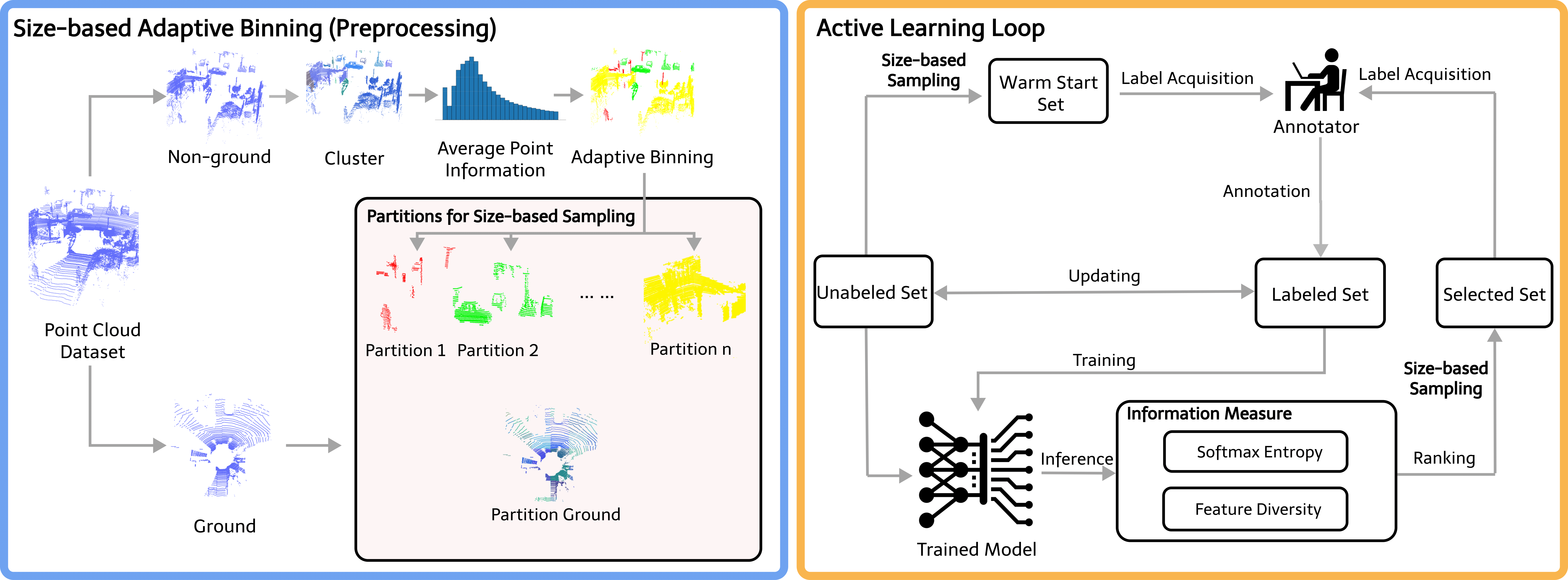}
    \caption{\textbf{Overview of BaSAL.}
    Our framework consists of a preprocessing step and the active learning loop.
    Size-balanced sampling is used to determine both the warm start set for the initialization of the active learning loop and the selected set for subsequent active learning iterations of the active learning loop.
    }
    \vspace{-5mm}
    \label{fig:overview}
\end{figure*}

\fig{overview} shows an overview of BaSAL, which consists of a preprocessing step and the active learning loop. 
We first preprocess the input point clouds to create multiple partitions (\sect{size_based_adaptive_binning}).
During the active learning loop, we use size-based sampling (\sect{size_based_sampling}) in both the model initialization phase (warm start) and subsequent active learning iterations of the loop to ensure a size-balanced training set. 
After warm starting the model, we adopt the established information measures (\sect{information_measure}) to rank and select the most informative clusters from each partition, until the annotation budget is reached.
Then we add them to the labeled set and retrain the model.

\subsection{Size-based Adaptive Binning (Preprocessing)}
\label{sec:size_based_adaptive_binning}
\begin{figure}[!htb]
    \centering
    \includegraphics[width=1.0\linewidth]{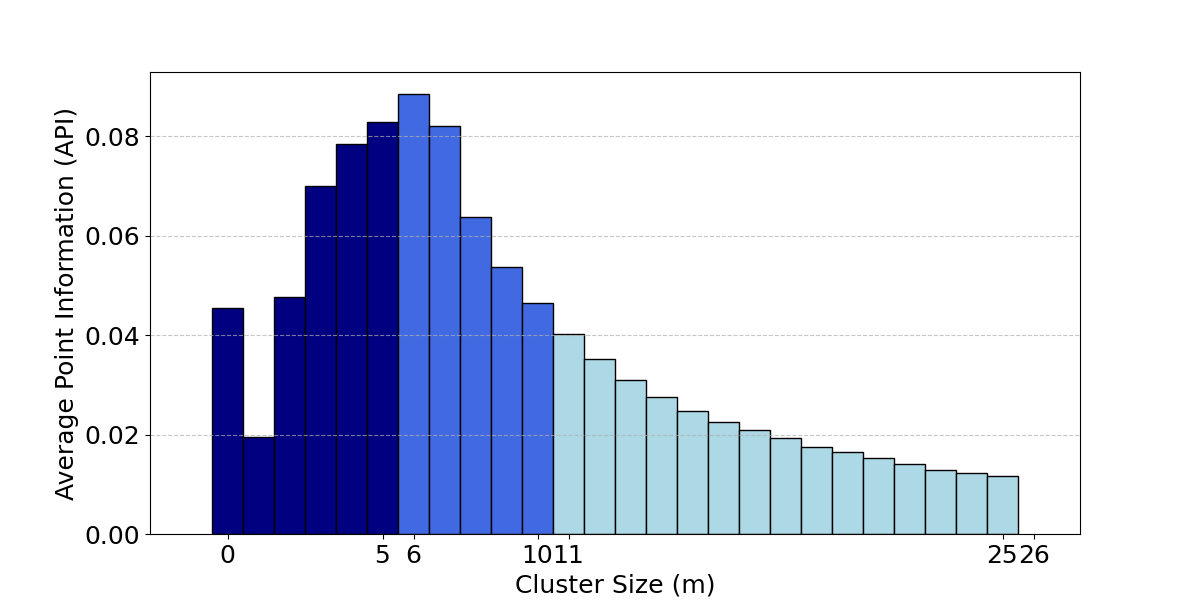}
    \caption{\textbf{Adaptive binning over sizes.} We split all size groups into 3 partitions, as indicated by colors. The accumulated API from each partition is approximately the same. We balance object classes by creating a size-based partition.}
    \label{fig:pid_bin}
\end{figure}

Given a dataset $\mathcal{D}$, we first apply ground plane detection and obtain the ground points which are further split into a number of grids $\mathcal{G}$. The remaining non-ground points are split into a collection of class-agnostic clusters $\mathcal{C}$ using HDBSCAN~\cite{hdbscan}. 
Based on the assumption that objects of similar sizes tend to share the same semantic labels, we group the class-agnostic clusters $\mathcal{C}$ by size, i.e. the sum of the dimensions of the 3D bounding box that encloses a cluster.
We round all sizes to integers. 
To balance the information among sizes, we calculate the Average Point Information (API) per size, defined by \eq{pid}, 
where $C_{s}$ is the number of clusters and $P_{s}$ is the number of accumulated points from all clusters, for a particular size $s$. 
$\frac{P_{s}}{C_{s}}$ stands for the average number of points per cluster and the additional logarithm penalizes larger clusters, as each additional point brings less information gain. 
We adaptively bin all sizes by API and obtain $B$ partitions such that the accumulated API inside each partition is approximately the same.
We show an example in \fig{pid_bin}, where $B$ is set to 3.
\begin{equation}
    API_{s} = \left(\frac{log P_{s}}{C_{s}}\right)^{-1}
    \label{eq:pid}
\end{equation}

\subsection{Size-based Sampling}
\label{sec:size_based_sampling}
We employ size-based sampling in two places in our active learning pipeline.
In the initial active learning iteration of the loop, most works~\cite{redal,lidal} use random sampling. 
Instead, we allocate the budget equally to all partitions (including the ground partition), and then randomly sample clusters from each partition.
This achieves the desired warm start.
In subsequent iterations, we rank clusters within each size-based partition using standard information measures (\sect{information_measure}) and sample the top-ranking clusters for annotation and further training. The budget remains approximately the same over partitions.

\vspace{-1mm}
\subsection{Information Measures}
\label{sec:information_measure}
To rank clusters, we combine two well-established information measures: Softmax entropy~\cite{entropy} and feature diversity (CoreSet)~\cite{coreset}. Entropy aims to select the data that the model is most uncertain of. CoreSet prioritizes unlabeled data that is far from the labeled data in the feature space to ensure diversity. 
After ranking, we choose the top-ranking clusters for training in the next active learning iteration.

\subsubsection{Softmax Entropy}
We calculate the entropy for each unlabeled cluster $j$ via \eq{softmax_entropy}, where $N$ is the number of points in cluster $j$ and $p_{n,c}$ is the predicted probability of point $n$ and class $c$.
\begin{equation}
    E_j= - \frac{1}{N} \sum_{n=1}^{N} \sum_{c=1}^{C} p_{n,c} \log(p_{n,c})
\label{eq:softmax_entropy}
\end{equation}

\begin{figure*}
    \centering
    \includegraphics[width=1.0
\textwidth]{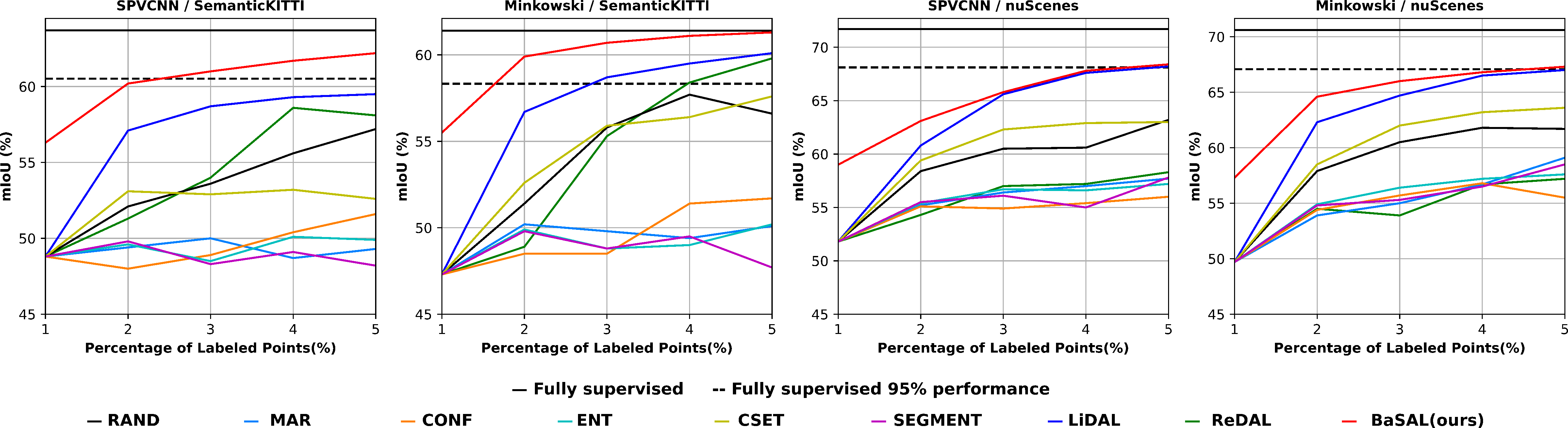}
    \caption{\textbf{Experiment results of different active learning strategies on SemanticKITTI~\cite{SemanticKitti}, nuScenes~\cite{nuScene} using SPVCNN~\cite{searching}, Minkowski~\cite{minkowski} network.} 
    We compare BaSAL with other existing works. 
    The solid line is the performance of the fully supervised model.
    The dashed line indicates 95\% performance of the fully supervised model.
    Our model outperforms all existing active learning approaches on SemanticKITTI and gets on par performance with the state-of-the-art active learning method LiDAL~\cite{lidal} on nuScenes.}
    \label{fig:experiments}
\end{figure*}

\subsubsection{Feature Diversity}
We use the feature from the last layer in the encoder-decoder network before classification and calculate the mean feature vector $\textbf{f}$ for each cluster.
The feature diversity $D_{j}$ is defined as the summed Euclidean distance between an unlabeled cluster $j$ and all labeled clusters from the same partition, as in \eq{feature_diversity}, where $i$ is the index of a labeled cluster.
\begin{equation}
\vspace{-1mm}
\begin{aligned}
    D_{j} &= \sum_{i=1}^{N_l}||\textbf{f}_i - \textbf{f}_j||_2 
\end{aligned}
\label{eq:feature_diversity}
\end{equation}

\subsubsection{Combination}
To create the final ranking of unlabeled clusters, we sort all unlabeled clusters by entropy and diversity per partition, as defined in~\eq{ud_combination}, where $r_{E_j}$ and $r_{D_j}$ denote the ranks for the $j$-th cluster in terms of entropy and diversity. 
$\mathcal{U}_{b}$ contains the indices of all unlabeled clusters in partition $b$, and $R_{b}$ is the overall ranking of these clusters. 

\begin{equation}
\begin{aligned}
    R_{b} = \underset{j \in \mathcal{U}_{b}}{argsort} \left(\frac{1}{r_{E_{j}}} + \frac{1}{r_{D_{j}}}\right)
\end{aligned}
\label{eq:ud_combination}
\end{equation}
\section{Experiments}
We conduct extensive experiments on SemanticKITTI~\cite{kitti} and nuScenes~\cite{nuScene} datasets, and compare with baselines~\cite{lidal, redal} on LiDAR semantic segmentation task.

\subsection{Datasets and Evaluation Metric}
On SemanticKITTI, we take sequences $00 - 07$ and seq $09 - 10$ for training and report the performance on validation sequence $08$.
On nuScenes, we train our model on the official training split which contains 700 scenes, and report the performance on the validation split with 150 scenes. We compare the mean Intersection over Union (mIoU) metric over 19 classes in SemanticKITTI and 16 classes in nuScenes.

\subsection{Experimental Settings}
\subsubsection{Network Architectures}
Following ReDAL~\cite{redal} and LiDAL~\cite{lidal}, we test our model with two backbone architectures: SPVCNN~\cite{searching} and MinkowskiNet~\cite{minkowski}.

\subsubsection{Baselines}
\label{sec:baseline al methods}
We compare with random point selection (RAND), softmax confidence (CONF)~\cite{softmax}, softmax margin (MAR)~\cite{softmax}, softmax entropy (ENT)~\cite{softmax}, MC-dropout (MCDR)~\cite{dropout}, CoreSet selection (CSET)~\cite{coreset}, segment-entropy (SEGENT)~\cite{efficient}, ReDAL~\cite{redal} and LiDAL~\cite{lidal}.
Experimental results of the baselines come from LiDAL~\cite{lidal}.

\begin{figure*}
     \centering
     \begin{subfigure}[b]{0.24\textwidth}
         \centering
         \includegraphics[width=\textwidth]{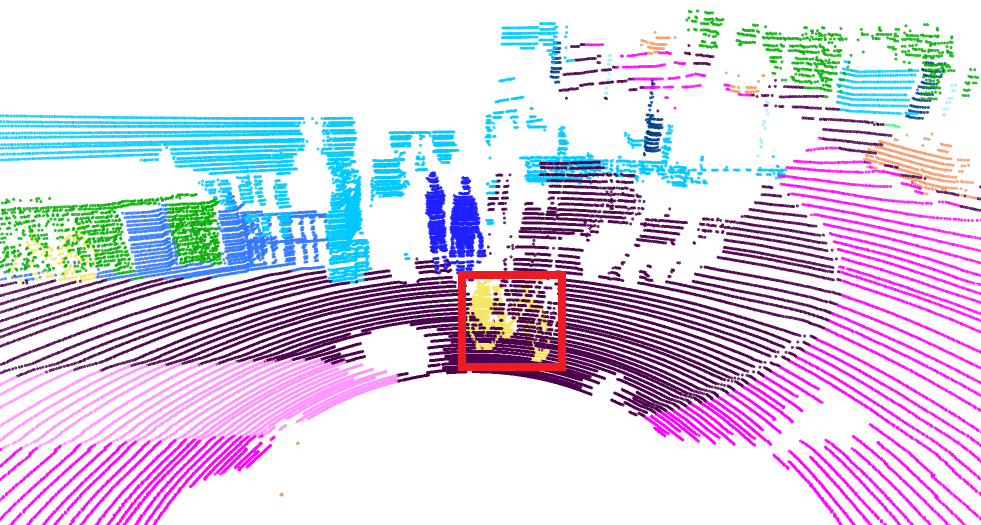}
         \caption{Ground Truth}
         \label{fig: label}
     \end{subfigure}
     \begin{subfigure}[b]{0.24\textwidth}
         \centering
         \includegraphics[width=\textwidth]{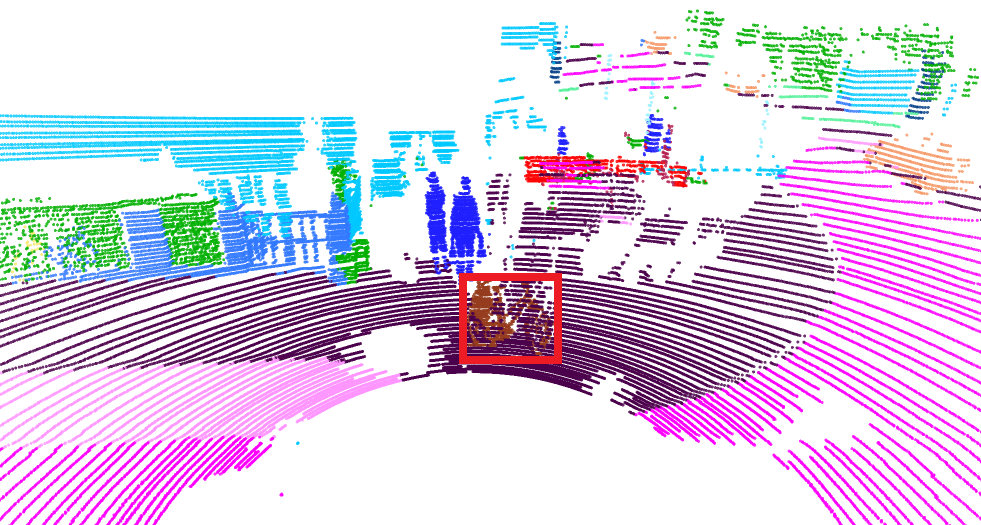}
         \caption{Fully Supervised (100\%)}
         \label{fig: fs}
     \end{subfigure}
     \begin{subfigure}[b]{0.24\textwidth}
         \centering
         \includegraphics[width=\textwidth]{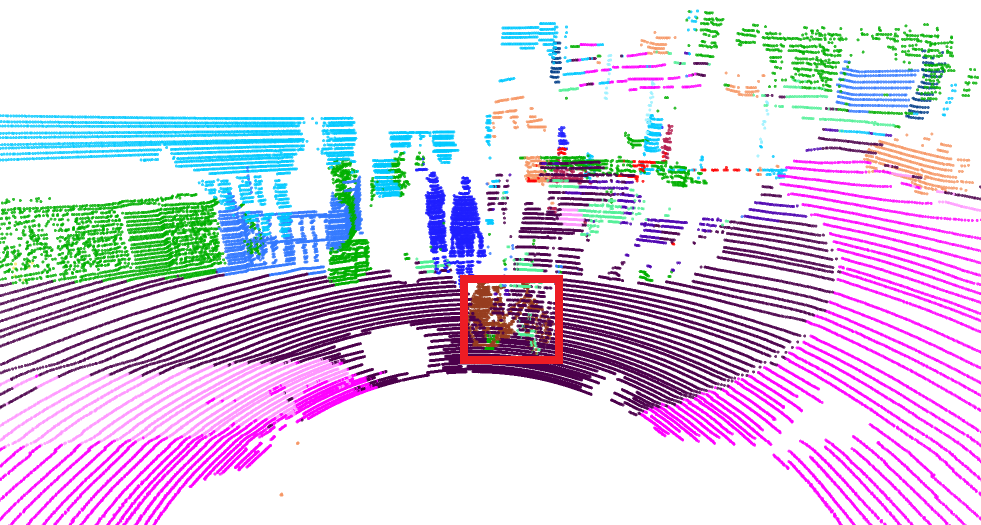}
         \caption{ReDAL (5\%)}
         \label{fig: redal}
     \end{subfigure}
     \begin{subfigure}[b]{0.24\textwidth}
         \centering
         \includegraphics[width=\textwidth]{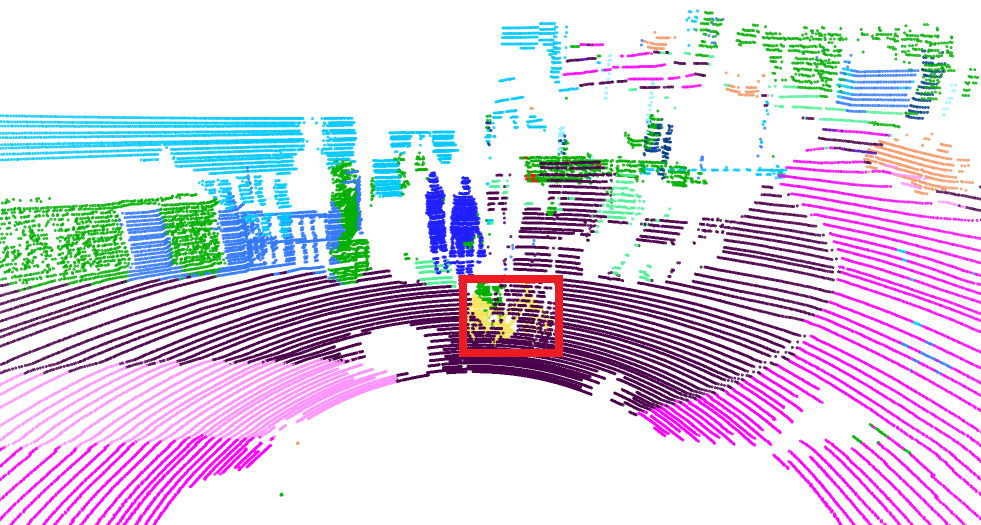}
         \caption{BaSAL (5\%)}
         \label{fig: basal}
     \end{subfigure}
     \begin{subfigure}[b]{0.24\textwidth}
         \centering
         \includegraphics[width=\textwidth]{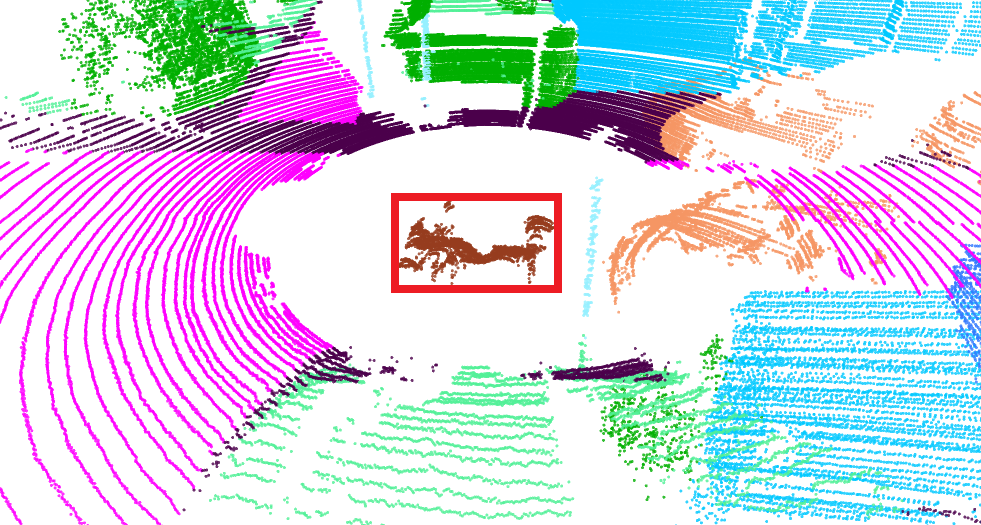}
         \caption{Ground Truth}
         \label{fig: label_2}
     \end{subfigure}
     \begin{subfigure}[b]{0.24\textwidth}
         \centering
         \includegraphics[width=\textwidth]{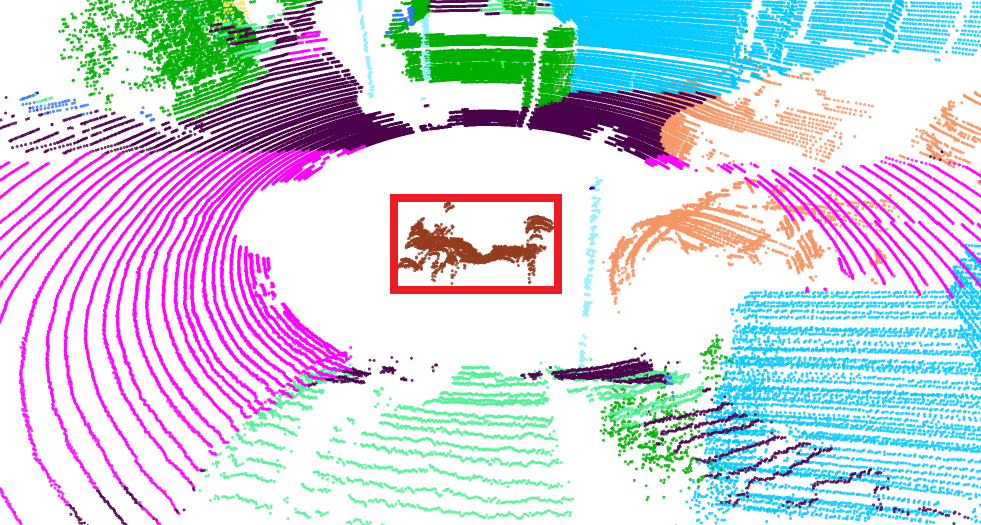}
         \caption{Fully Supervised (100\%)}
         \label{fig: fs_2}
     \end{subfigure}
     \begin{subfigure}[b]{0.24\textwidth}
         \centering
         \includegraphics[width=\textwidth]{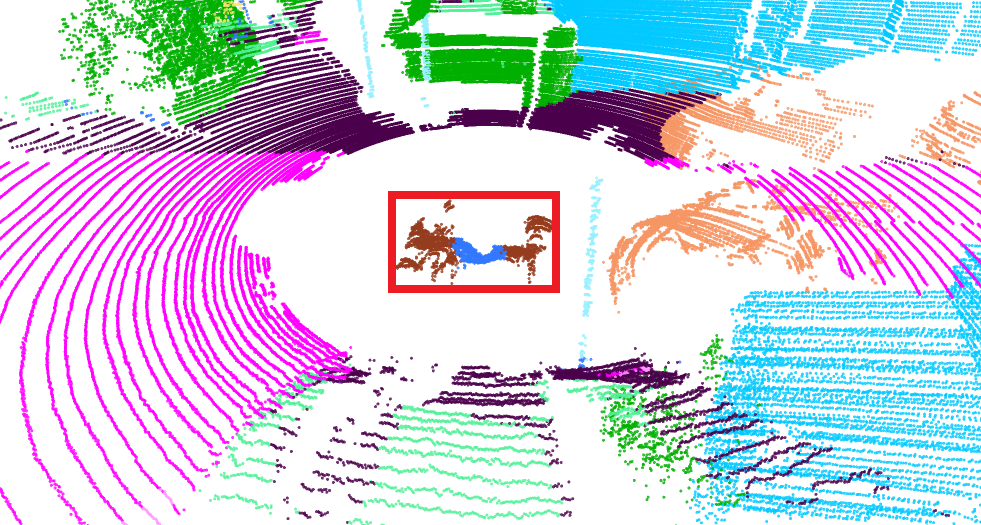}
         \caption{ReDAL (5\%)}
         \label{fig: redal_2}
     \end{subfigure}
     \begin{subfigure}[b]{0.24\textwidth}
         \centering
         \includegraphics[width=\textwidth]{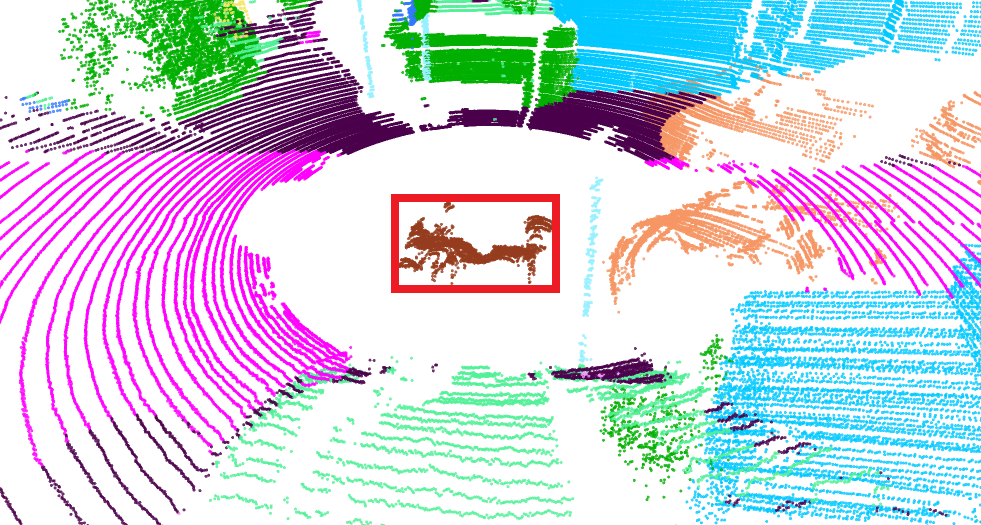}
         \caption{BaSAL (5\%)}
         \label{fig: BaSAL_2}
     \end{subfigure}
    \caption{\textbf{Qualitative comparison on SemanticKITTI~\cite{SemanticKitti} using the Minkowski~\cite{minkowski} backbone.} We visualize semantic segmentation results on two examples. Our model successfully detects the bicycle in (d), as indicated by the red box. In comparison, other models misclassify the bicycle as sidewalks in (b) and (c).  In the second example, our model better detects the motorcycle in (h), while ReDAL~\cite{redal} over-segments it.   
    We improve the performance on underrepresented classes. }
    \label{fig:Qualitative_results}
\end{figure*}

\subsubsection{Active Learning Protocol}
BaSAL training has two stages: warm start initialization and subsequent active learning iterations.
We warm start the model by size-based sampling. The annotation budget in this phase is $x_{init} \%$. After warm start, we conduct $K$ active learning iterations.
During each iteration, we interactively select top-ranking clusters from each partition for label acquisition such that the total annotation is no greater than $x_{active}\%$. Then we load the previous checkpoint and fine-tune the model with all labeled data.

The labeling budget is measured using the percentage of the labeled points with respect to the entire dataset. For both SemanticKITTI and nuScenes, we set $x_{init} = 1$, $K = 4$, and $x_{active} = 1$. The annotation budget is uniformly distributed over all partitions (including the ground partition) in each iteration. We report the average over three runs.

\subsection{Implementation Details}
\subsubsection{Network Training}
All experiments are conducted with a single A40 GPU.
For SemanticKITTI and nuScenes, we set the training batch size to 10/30 and validation batch size to 20/60.
We first train our network using the warm start data for 100/200 epochs.
Then we finetune the model for 30/150 epochs for each active learning iteration. 
For both datasets, we train the networks with the cross-entropy loss and the Adam optimizer with an initial learning rate of 1e-3. 

\subsubsection{Size-based Adaptive Binning}
Given a point cloud dataset, we first use PatchWork++~\cite{patchworkpp} with the default parameters to separate the ground points and the non-ground points. 
The ground points are then divided into grids of size 10m $\times$ 10m, named Partition Ground. 
For non-ground points, we use HDBSCAN clustering ~\cite{hdbscan} algorithm to split them into clusters. We set $min\_cluster$, $min\_samples$ and $cluster\_selection\_epsilon$ to be 20, 10 and 0.5 for SemanticKITTI and 20, 1 and 0.5 for nuScenes.
We then calculate the size of all the non-ground clusters. The size of a cluster is the rounded sum of the length, width, and height of its bounding box. 
We filter those clusters with the size larger than 25 meters. As mentioned in Section~\ref{sec:size_based_adaptive_binning}, we calculate the Average Point Information (API) metric and adaptively group the sizes to 3 partitions.

\subsubsection{Size-based Sampling}
The size-based sampling method is used in both the warm start step and subsequent active learning iterations. For warm start (1\% budget), we uniformly sample clusters from Partition 1, Partition 2, Partition 3, and Partition Ground until the accumulated number of points in each Partition reaches 0.25\% of the entire dataset. 
For each subsequent active learning iteration, we increase the point budget by 1\%, which is also uniformly allocated to all partitions (0.25\% each).

\subsection{Main Results}
\fig{experiments} compares BaSAL with the baselines. The x-axis represents the percentage of labeled points with respect to the entire training set, and the y-axis represents performance (mIoU). 
On SemanticKITTI, our model consistently outperforms all baselines in all settings.
When the annotation budget increases to $5\%$, BaSAL (with Minkowski backbone) is able to match the performance of fully supervised training on the entire dataset, demonstrating the benefit of active learning in reducing annotation effort.
BaSAL with the SPVCNN backbone achieves 98$\%$ the performance of the fully supervised model.
On nuScenes, BaSAL surpasses all the baselines given $1\%$, $2\%$, and $3\%$ labeled data. 
LiDAL\cite{lidal} is able to match the performance of BaSAL when the annotation budget is up to $5\%$. Both models reach $95\%$ performance of fully supervised learning.
We also compare our warm start with the random sampling cold start at $1\%$.
Notably, the advantage of our model over other baselines is more than $5\%$ mIoU on both datasets, validating the advantage of our warm start in tackling the cold start problem. 

\begin{table*}[ht]
    \caption{\textbf{Class distribution on SemanticKITTI~\cite{SemanticKitti}.}
    We calculate the class distribution over the entire dataset and over partitions of different size.
    There are 4 partitions (including the ground partition). 
    Classes are highly imbalanced over the dataset and most classes have a characteristic size (highlighted in bold text).
    }
    \centering
    \resizebox{\textwidth}{!}{\begin{tabular}{c|c c c c c c c c c c c c c c c c c c c }
        \hline
         &car &bicycle &motorcycle &truck & \makecell{ other \\ vehicle } &person &bicyclist &motorcyclist &road &parking &sidewalk &\makecell{other \\ ground} &building &fence &vegetation &trunk &terrain &pole & \makecell{traffic \\sign} \\
        \hline
            Entire dataset &5.7 &0.01 &0.05 &0.3 &0.3 &0.04 &0.02 &0.005 & 27.3 &2.0 &19.7 &0.5 &10.2 &3.9 &18.6 &0.8 &10.3 &0.3 &0.07 \\
        \hline
        Partition1 (small) &26.7 &23.4 &\textbf{77.3} &3.6 &17.7 &\textbf{53.9} &\textbf{68.2} &\textbf{76.3} &0.1 &0.1 &0.2 &1.2 &6.5 &4.4 &9.3 &\textbf{48.7} &1.1 &\textbf{42.9} &\textbf{50.5} \\
        Partition2 (medium) &\textbf{63.9} &14.3 &10.6 &29.6 &\textbf{56.5} &23.2 &24.3 &18.1 &0.1 &0.2 &0.2 &1.6 &13.1 &9.0 &18.0 &33.2 &1.7 &19.6 &23.6 \\
        Partition3 (large) &6.5 &\textbf{40.4} &6.9 &\textbf{62.9} &23.7 &17.8 &5.6 &0.8 &0.1 &0.1 &0.4 &4.4 &\textbf{71.1} &\textbf{49.8} &\textbf{50.7} &14.7 &3.4 &31.6 &25.7 \\
        Ground Partition &2.8 &21.9 &5.2 &3.9 &2.1 &5.1 &1.9 &4.9 &\textbf{99.7} &\textbf{99.6} &\textbf{99.3} &\textbf{92.7} &9.3 &36.8 &22.1 &3.5 &\textbf{93.9} &5.8 &0.2 \\
        \hline
    \end{tabular}}

\label{tab:class_distribution}
\end{table*}

\begin{table}    \caption{\textbf{Performance on less frequent classes.} We measure mIoU (\%) on SemanticKITTI~\cite{SemanticKitti} using the Minkowski~\cite{minkowski} backbone. We also calculate the proportion of points per class with respect to the full dataset. Our model boosts the performance on rare classes substantially, verifying its advantage in tackling class imbalance. FS stands for fully supervised learning on the full dataset.}
    {\begin{tabular}{c|c c c c c}
        \hline
         &bicycle &motorcycle &person &bicyclist &motorcyclist\\
        \hline
        $\%$ &0.01 &0.05 &0.04 &0.02 &0.005\\
        \hline
        RAND &9.5 &45.0 &52.0 &47.8 &0.0\\
        ReDAL &29.6 &58.6 &63.4 &84.1 &\textbf{0.5}\\
        FS &20.4 &63.9 &65.0 &78.5 &0.4 \\
        Ours &\textbf{43.5} &\textbf{70.6} &\textbf{70.4} &\textbf{88.4} &0.2\\
        \hline
    \end{tabular}}
    \label{tab:result_per_class}
\end{table}

\subsection{Ablation Study}

\begin{table}[ht]
    \caption{\textbf{Ablation study.} We conduct our ablation on SemanticKITTI~\cite{SemanticKitti}. We train all models from scratch in iteration 1 and fine-tune them in iteration 2-5. Our size-based partition contributes the most to the final performance.  RS: random sampling from all unlabeled data; SS: size-based sampling; EN: softmax entropy as the information measure; FD: feature diversity as the information measure~\cite{coreset}. }
    \centering
    {\begin{tabular}{c|cc|cc|c}
        \hline
          Iteration 1 & \multicolumn{4}{c|}{Iteration 2-5}&  \multirow{3}{*}{mIoU(\%)}\\
        \cline{1-5}
          \multirow{2}{*}{Initialization}& \multicolumn{2}{c|}{Info measure} & \multicolumn{2}{c|}{Data Sampling}&  \multirow{2}{*}{}\\
        \cline{2-5}
        & EN & FD &RS & SS & \\
        \hline
       \multirow{2}{*}{\makecell{Cold \\start}} &   &  &\checkmark  &  &56.8\\
            
        &  &   &  & \checkmark &60.1\\
             \hline
        \multirow{3}{*}{\makecell{Warm \\start}}&    &  &  & \checkmark &60.4\\
      
         & \checkmark &  &  & \checkmark &61.1 \\
       
         & \checkmark&\checkmark  &   & \checkmark  &61.3\\
        \hline
    \end{tabular}}
    \label{tab:ablation}
\end{table}

\subsubsection{Method Ablation}
We numerically evaluate the contribution of each component in our design on SemanticKITTI~\cite{SemanticKitti} in \tab{ablation}.
RS stands for random sampling from all unlabeled clusters. SS represents size-based sampling. EN and FD indicate the information measures: entropy and feature diversity, respectively. 
We first ablate our size-based partition.
Without relying on any learning-based information measures, our model gains a 3.3\% improvement, by simply replacing the random sampling with our size-based partition in the cold start setting. Notably, this is the most significant increase in our ablation study.
Warm starting the model further results in a $0.3\%$ improvement. 
Next, we evaluate the effect of information measures. Adding softmax entropy as an information measure improves the result by $0.7\%$. Taking feature diversity as an extra measure marginally improves the result by $0.2\%$.

\subsubsection{Number of Partitions}
We also experiment with different numbers of partitions when warm starting the model. We set the number of partitions $B$ to 3, 6, 12, and 25, and achieve the best performance when $B$ is set to 3 and 6. When $B$ gets larger to 12 and 25, the performance drops. In general, we find the number of partitions has a minor impact on the performance when $B$ is between 3 and 6. We set $B$ to 3 in our implementation.

\subsection{Analysis of Class Imbalance}
\subsubsection{Analysis on Less Frequent Classes}
We show our results on less frequent classes in \tab{result_per_class}.  
We outperform ReDAL and fully supervised learning by a substantial margin on four classes: bicycle, motorcycle, person and bicyclist. For example, the advantage over baselines is more than $20\%$ on the bicycle class and approximately $10\%$ on the bicyclist class. It is worth noting that these two classes only occupy $0.01\%$ and $0.02\%$ of the entire dataset. We attribute the remarkable improvement to our size-based partition which better handles class-balance.
We show qualitative results on two samples in \fig{Qualitative_results}.  In the top row, our model is able to recognize the bicycle, labeled by the red box in (d), while other models fail. In the second row, our model better segments the motorcycle than ReDAL~\cite{redal}, verifying our advantage in segmenting underrepresented classes.

\subsubsection{Analysis on Size-based Partition}
We also analyze the distribution of all classes on SemanticKITTI in \tab{class_distribution}.
We first enumerate the proportion of each class and then calculate the distribution of each class after size-based partition. All the numbers are vertically normalized over partitions per class. One key observation is that each object class has a size bias, which aligns with our assumption that size is an informative cue to distinguish classes. For example, the less frequent classes, \textit{person}, \textit{motorcyclist} and \textit{motorcycle}, come mostly from \textit{partition 1}. 
Compared to random sampling from the entire dataset, sampling from \textit{partition 1} is likely to increase the chance of obtaining an underrepresented object class, thus reducing the class imbalance considerably. 
However, we also notice that the majority of the \textit{bicycle} class falls in \textit{partition 3}. We suspect that this class may not be well segmented by the HDBSCAN ~\cite{hdbscan} clustering algorithm. 
\section{Discussion} 
Active learning in large-scale autonomous driving datasets aims to reduce the heavy labeling effort.
However, recent works quantify labeling effort differently.
\cite{exploring} counts the number of frames and bounding boxes.
ReDAL~\cite{redal}, LiDAL~\cite{lidal}, Just Label What You Need~\cite{justlabel} count the number of point labels.
Some other works measure the number of clicks of an annotator, such as OneThingOneClick~\cite{onething} and LESS~\cite{less}.
LESS achieves approximately 100\% performance using only 0.1\% of the annotations.
A crucial difference between our method and LESS is that they require the annotator to inspect the entire dataset, whereas in our case the annotator only inspects 5\% of the data.

\section{Conclusion}
We present BaSAL, a size-based warm start strategy for active learning on LiDAR semantic segmentation. BaSAL reduces the class imbalance in large-scale autonomous driving datasets by sampling from a size-based partition.
Our intuition is that each object class has a characteristic size.
Thus we are able to indirectly address the class imbalance and the cold start problem by grouping class-agnostic clusters according to size. 
We fuse the size-based partition into state-of-the-art active learning models and improve the performance considerably.
Particularly, on SemanticKITTI we achieve the same performance as fully supervised learning on the entire dataset, while using only $5\%$ of the annotation. Meanwhile, we boost the performance on less frequent classes significantly. 
Future work can explore better ways to create object clusters and develop novel cost functions to accurately quantify human annotation effort. On-device active learning is also an interesting direction without an excessive demand for computational resources.

\bibliographystyle{./IEEEtran} 
\bibliography{reference}
\clearpage
\section*{\large{\textbf{Supplementary Material}}}

The supplementary material is organized as follows:  
Section~\hyperref[supp: cluster]{A} visualizes the query units adopted in baselines and our work. 
Section~\hyperref[supp: experiment results]{B} provides more details for the experimental results on SemanticKITTI~\cite{SemanticKitti} and nuScenes~\cite{nuScene}.
\phantomsection

\subsection*{A. Visual comparison of query units}
\label{supp: cluster}
\begin{figure}[h]
     \centering
     \begin{subfigure}[b]{0.45\textwidth}
         \centering
         \includegraphics[width=\textwidth]{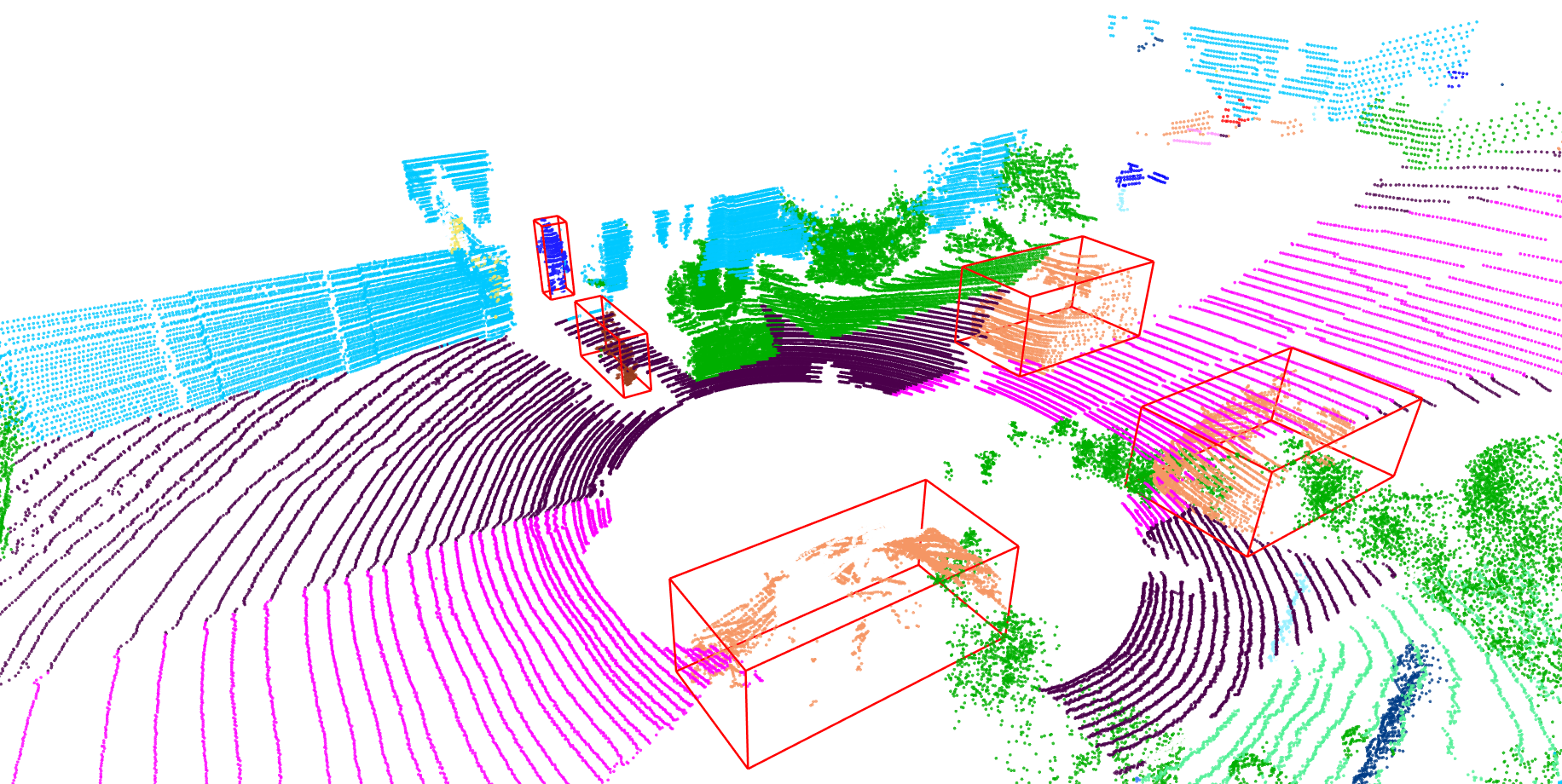}
         \caption{Ground truth semantic labels.}
         \label{fig:gt}
     \end{subfigure}
     \begin{subfigure}[b]{0.45\textwidth}
         \centering
         \includegraphics[width=\textwidth]{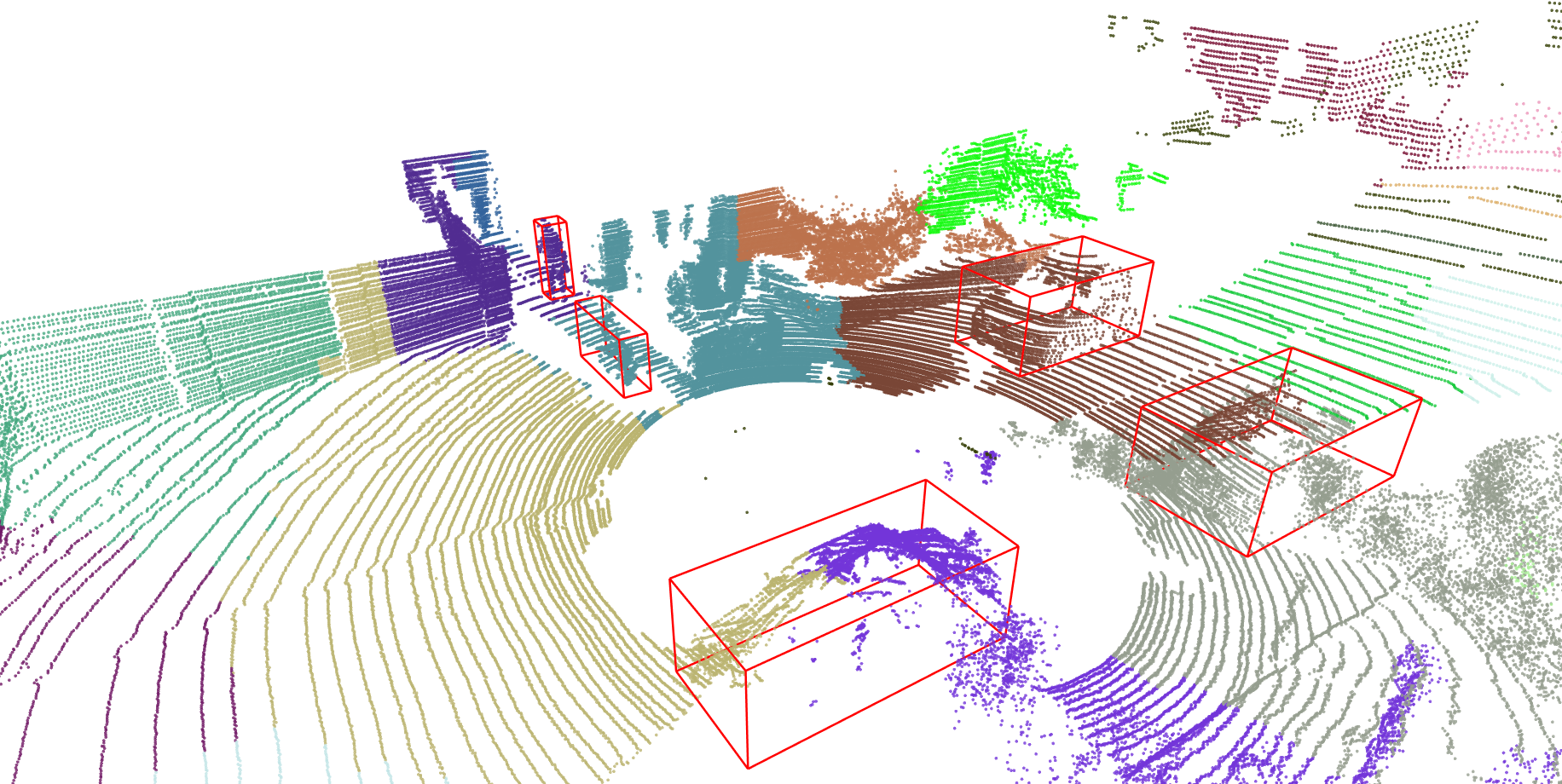}
         \caption{Supervoxels used in ReDAL \cite{redal} and LiDAL \cite{lidal}.}
         \label{fig:supervoxel}
     \end{subfigure}
     \begin{subfigure}[b]{0.45\textwidth}
         \centering
         \includegraphics[width=\textwidth]{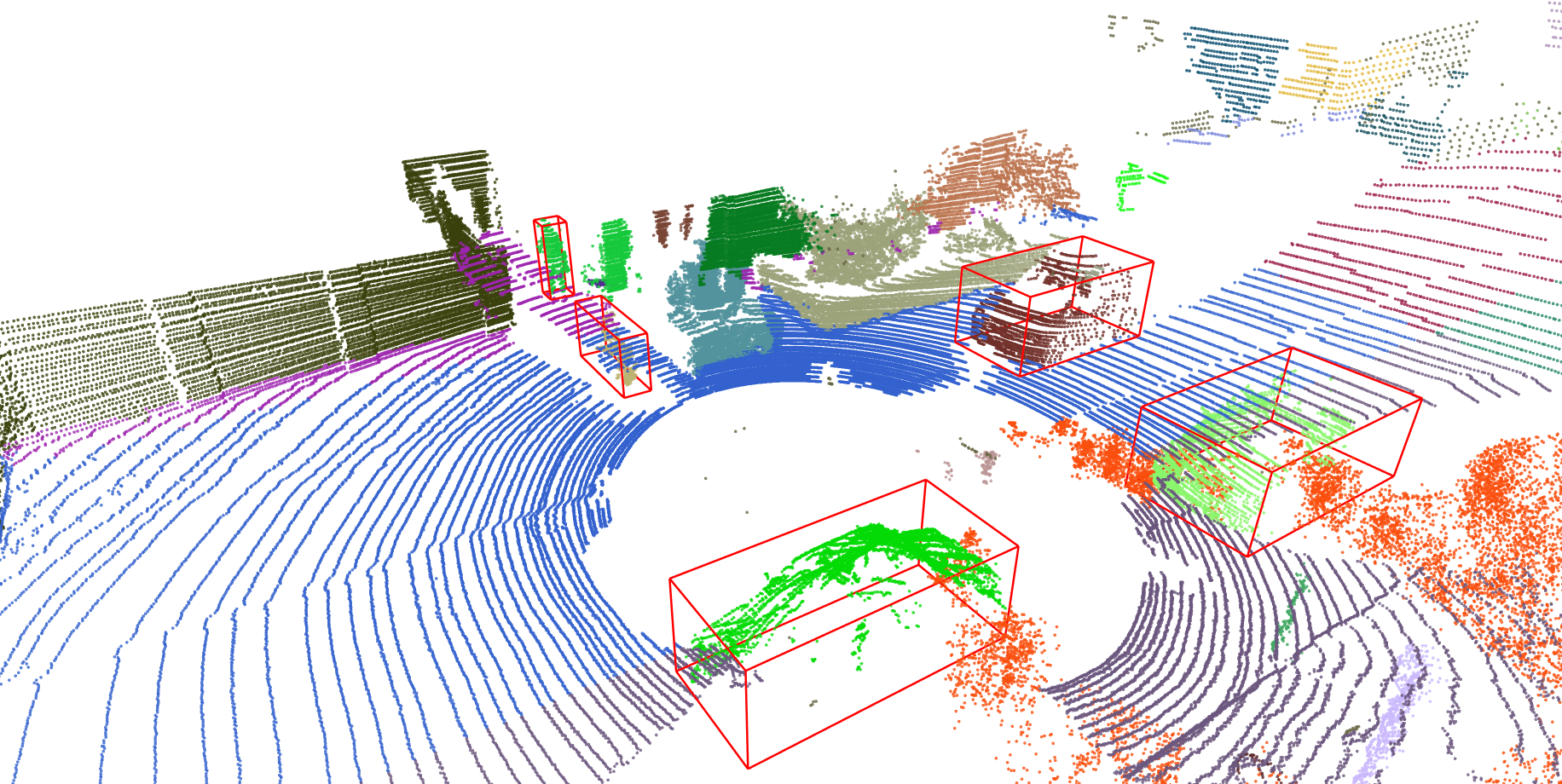}
         \caption{Object clusters used in our work.}
         \label{fig:cluster}
     \end{subfigure}
    \caption{\textbf{Visualization of the object clusters and the supervoxels.} Points with the same color belong to one supervoxel (cluster). Red bounding boxes indicate ground truth object clusters.}
    \label{fig:Supp_cluster}
\end{figure}

Different active learning approaches query the oracle annotator to label samples at different levels of granularity, which we call \emph{query units}.
Common query units include entire point clouds~\cite{exploring,justlabel}, supervoxels ~\cite{redal,lidal} or the more fine-grained object clusters used in our model.
We qualitatively compare the supervoxels and object clusters in \fig{Supp_cluster}. 
ReDAL and LiDAL use the VCCS~\cite{vccs} algorithm to construct supervoxels.
As shown in \fig{supervoxel}, the algorithm separates the point cloud frame to several connected supervoxels. 
However, compared with the ground truth object clusters, the supervoxels of the baselines are coarse, which often puts many different objects in one supervoxel (undersegmentation). 
In addition, the car at the bottom is segmented into two different supervoxels (oversegmentation). 
In contrast, as shown in \fig{cluster}, our clustering pipeline separates different objects well, making the querying and sampling process more fine-grained and accurate.

\subsection*{B. Details on the Main Results}
\label{supp: experiment results}
The baseline experiment results are taken from LiDAL~\cite{lidal}, including random selection (RAND), softmax confidence (CONF), softmax margin (MAR), softmax entropy (ENT), MC-Dropout (MCDR), Core-Set selection (CSET), segment-entropy (SEGMENT), ReDAL, and LiDAL.

\tab{table1} shows the numerical results (mIoU) on the SemanticKITTI validation set with the SPVCNN \cite{searching} network, which aligns with \fig{experiments}. 
With 1\% labeling budget, we achieve 56.3\% mIoU, outperforming the baselines (48.8\%) by a large margin.
When the budget increases to 5\%, we reach 62.2\% mIoU, improving previous state-of-the-art by 2.7\%.
\begin{table}[H]
    \caption{mIoU (\%) on SemanticKITTI~\cite{SemanticKitti} validation set with SPVCNN~\cite{searching} network.
    }
    \resizebox{0.47\textwidth}{!}{\begin{tabular}{c|c|c|c|c|c}
        \hline
        Methods &Init 1\% & 2\% & 3\% & 4\% & 5\%\\
        \hline
        RAND &48.8 & 52.1 & 53.6 & 55.6 & 57.2\\
        \hline
        MAR  &48.8 & 49.4 & 50.0 & 48.7 & 49.3\\
        \hline
        CONF &48.8 & 48.0 & 48.9 & 50.4 & 51.6\\
        \hline
        ENT  &48.8 & 49.6 & 48.5 & 50.1 & 49.9\\
        \hline
        CSET &48.8 & 53.1 & 52.9 & 53.2 & 52.6\\
        \hline
        SEGMENT &48.8 & 49.8 & 48.3 & 49.1 & 48.2\\
        \hline
        ReDAL &48.8 & 51.3 & 54.0 & 58.6 & 58.1\\
        \hline
        LiDAL &48.8 & 57.1 & 58.7 & 59.3 & 59.5\\
        \hline
        BaSAL (Ours) &\textbf{56.3} & \textbf{60.2} & \textbf{61.0} & \textbf{61.7} & \textbf{62.2}\\
        \hline
    \end{tabular}}

    \label{tab:table1}
\end{table}

\tab{table2} shows the mIoU results on SemanticKITTI validation set with Minkowski~\cite{minkowski} network. We achieve an 8.2\% improvement over other baselines using 1\% annotation budget. 
Given 5\% annotation budget, we reach 61.3\% mIoU, outperforming the previous state-of-the-art.
\begin{table}[H]
    \caption{mIoU (\%) on SemanticKITTI~\cite{SemanticKitti} validation set with Minkowski~\cite{minkowski} network.
    }
    \resizebox{0.47\textwidth}{!}{\begin{tabular}{c|c|c|c|c|c}
        \hline
        Methods &Init 1\% & 2\% & 3\% & 4\% & 5\%\\
        \hline
        RAND &47.3 & 51.4 & 55.8 & 57.7 & 56.6\\
        \hline
        MAR  &47.3 & 50.2 & 49.8 & 49.4 & 50.1\\
        \hline
        CONF &47.3 & 48.5 & 48.5 & 51.4 & 51.7\\
        \hline
        ENT  &47.3 & 49.9 & 48.8 & 49.0 & 50.2\\
        \hline
        CSET &47.3 & 52.6 & 55.9 & 56.4 & 57.6\\
        \hline
        SEGMENT &47.3 & 49.8 & 48.8 & 49.5 & 47.7\\
        \hline
        ReDAL &47.3 & 56.7 & 58.7 & 59.5 & 60.1\\
        \hline
        LiDAL &47.3 & 51.4 & 55.8 & 57.7 & 56.6\\
        \hline
        BaSAL (Ours) &\textbf{55.5} & \textbf{59.9} & \textbf{60.7} & \textbf{61}\textbf{.1} & \textbf{61.3}\\
        \hline
    \end{tabular}}

    \label{tab:table2}
\end{table}

\tab{table3} and \tab{table4} show the mIoU results on the nuScenes validation set with SPVCNN~\cite{searching} and Minkowski~\cite{minkowski} network, respectively. We substantially improve over the baselines by approximately 8\% using 1\% of the annotation budget. 
We also match the state-of-the-art when the budget increases to 5\%.
\begin{table}[H]
    \caption{mIoU (\%) on nuScenes~\cite{nuScene} validation set with SPVCNN~\cite{searching} network.}
    \resizebox{0.47\textwidth}{!}{\begin{tabular}{c|c|c|c|c|c}
        \hline
        Methods &Init 1\% & 2\% & 3\% & 4\% & 5\%\\
        \hline
        RAND &51.8 & 58.4 & 60.5 & 60.6 & 63.2\\
        \hline
        MAR  &51.8 & 55.2 & 56.4 & 57.0 & 57.7\\
        \hline
        CONF &51.8 & 55.1 & 54.9 & 55.4 & 56.0\\
        \hline
        ENT  &51.8 & 55.4 & 56.7 & 56.6 & 57.2\\
        \hline
        CSET &51.8 & 59.4 & 62.3 & 62.9 & 63.0\\
        \hline
        SEGMENT &51.8 & 55.5 & 56.1 & 55.0 & 57.8\\
        \hline
        ReDAL &51.8 & 54.3 & 57.0 & 57.2 & 58.3\\
        \hline
        LiDAL &51.8 & 60.8 & 65.6 & 67.6 & 68.2\\
        \hline 
        BaSAL (Ours) &\textbf{59.0} & \textbf{63.1} & \textbf{65.8} & \textbf{67.8} & \textbf{68.4}\\
        \hline
    \end{tabular}}
    \label{tab:table3}
\end{table}

\begin{table}[H]
    \caption{mIoU (\%) on nuScenes~\cite{nuScene} validation set with Minkowski~\cite{minkowski} network.                      
    }
    \resizebox{0.47\textwidth}{!}{\begin{tabular}{c|c|c|c|c|c}
        \hline
        Methods &Init 1\% & 2\% & 3\% & 4\% & 5\%\\
        \hline
        RAND &49.7 & 57.9 & 60.5 & 61.8 & 61.7\\
        \hline
        MAR  &49.7 & 53.9 & 55.0 & 56.7 & 59.1\\
        \hline
        CONF &49.7 & 54.4 & 55.7 & 56.8 & 55.5\\
        \hline
        ENT  &49.7 & 54.9 & 56.4 & 57.2 & 57.6\\
        \hline
        CSET &49.7 & 58.5 & 62.0 & 63.2 & 63.6\\
        \hline
        SEGMENT &49.7 & 54.8 & 55.3 & 56.5 & 58.5\\
        \hline
        ReDAL &49.7 & 54.5 & 53.9 & 56.7 & 57.2\\
        \hline
        LiDAL &49.7 & 62.3 & 64.7 & 66.5 & 67.0\\
        \hline
        BaSAL (Ours) &\textbf{57.3} & \textbf{64.6} & \textbf{66.0} & \textbf{66.8} & \textbf{67.3}\\
        \hline
    \end{tabular}}
    \label{tab:table4}
\end{table}
\end{document}